\documentclass[review]{elsarticle}

\usepackage{lineno,hyperref}

\modulolinenumbers[5]

\journal{Pattern Recognition}

\usepackage{multirow}
\usepackage{colortbl}
\usepackage{booktabs}
\newcommand{\etal}{~\textit{et al.}}

\usepackage{amssymb}
\usepackage{makecell}
\usepackage[ruled,vlined,linesnumbered]{algorithm2e}
\usepackage{subfigure}
\usepackage{multirow}
\newcommand{\nonl}{\renewcommand{\nl}{\let\nl\oldnl}}
\usepackage{graphicx}
\usepackage{textcomp}
\usepackage{bm} 
\usepackage{amsmath,amssymb,amsfonts}
\usepackage[section]{placeins}
\usepackage{caption}

\usepackage[dvipsnames,table,xcdraw]{xcolor}

\journal{Pattern Recognition}

\begin{document}

\begin{frontmatter}

\title{LCReg: Long-Tailed Image Classification with Latent Categories based Recognition}

\author[Infocomm]{Weide Liu\corref{cor1}}
\address[Infocomm] {Institute for Infocomm Research, A*STAR, Singapore 138632}

\author[NTU]{Zhonghua Wu\corref{cor1} }
\address[NTU] {Nanyang Technological University (NTU), Singapore 639798}

\author[NTU]{Yiming Wang}
\author[NTU]{Henghui Ding}
\author[Infocomm]{Fayao Liu}
\author[Infocomm]{Jie Lin}
\author[NTU]{Guosheng Lin \corref{cor2}} 

\cortext[cor1]{ indicates equal contribution.}
\cortext[cor2]{ Corresponding author. G. Lin is with School of Computer Science and Engineering, Nanyang Technological University (NTU), Singapore 639798 (e-mail: gslin@ntu.edu.sg).}

\begin{abstract}
In this work, we tackle the challenging problem of long-tailed image recognition. Previous long-tailed recognition approaches mainly focus on data augmentation or re-balancing strategies for the tail classes to give them more attention during model training. However, these methods are limited by the small number of training images for the tail classes, which results in poor feature representations.
To address this issue, we propose the Latent Categories based long-tail Recognition (LCReg) method. Our hypothesis is that common latent features shared by head and tail classes can be used to improve feature representation. Specifically, we learn a set of class-agnostic latent features shared by both head and tail classes, and then use semantic data augmentation on the latent features to implicitly increase the diversity of the training sample.
We conduct extensive experiments on five long-tailed image recognition datasets, and the results show that our proposed method significantly improves the baselines.
\end{abstract}

\end{frontmatter}
\section{Introduction}
With the successful development of Convolution Neural Networks (CNNs), image recognition has achieved great success on the ideally collected balanced datasets such as ImageNet.
However, real-world applications often involve natural image data that follows a long-tail distribution, in which a small number of classes have a large number of labeled images while most classes have only a few instances or annotations. The traditional fully supervised training strategy is not effective for these unbalanced datasets, which resulting the classification performance of the tail classes dropping quickly.

Long-tailed image recognition has been proposed to address the imbalanced training data problem. The main challenges are the difficulties of handling the small-data learning problems and the extremely imbalanced classification over all the classes. Most of the long-tailed recognition methods focus on generating more data samples of tail classes via data augmentation or using the re-balancing strategy to provide higher importance weights for the tail classes. For example, widely used data augmentation techniques like cropping, flipping, and mirroring are used to generate more data samples of the tail classes during the model training. However, we argue that the diversity of the training samples for the tail classes is still inherently limited due to the limited number of training images, which leads to subtle performance improvement for the long-tailed recognition task by using these conventional data augmentation methods.

Different from the conventional data augmentation methods, semantic data augmentation~\cite{isda} tries to augment the image features by adding class-aware perturbations. These perturbations are drawn from a multivariate normal distribution, with class-wise covariance matrices calculated from all available training samples. 
However, directly applying semantic data augmentation to the long-tailed recognition task may not be optimal, as the calculated covariance matrix for tail classes may not provide sufficient meaningful semantic directions for augmentation due to the limited number of training samples.
MetaSAug~\cite{metasaug} tries to solve the issue of imbalanced statistics by updating the class-wise covariance matrix through the minimization of the LDAM loss on the validation sets. However, the performance of this approach is still limited by the limited diversity and number of training samples for tail classes.

\begin{figure*}[t]
\centering
    \includegraphics[width=1\linewidth]{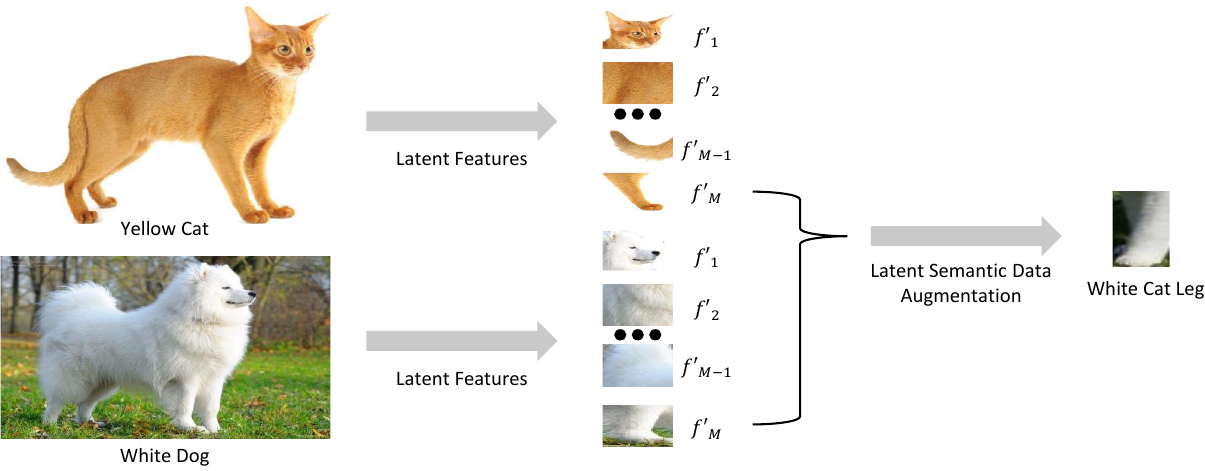}
    \caption{Our LCReg first projects the image features into the latent category features which share the commonality, such as the legs of cats and dogs. By performing the class semantic transformations along with the latent category, we aim to enrich the cat's feature by leveraging the common features, e.g., change the yellow cat leg by leveraging the dog's leg features. 
}
    \label{Fig:first}
\end{figure*}

To overcome the limitations mentioned above, we propose to mine out and augment the common features among the head and tail classes to increase the diversity of the training samples. The commonality is obtained with the assumption that objects from the same domain might share some commonalities. For instance, cats and dogs share a commonality of legs with similar shapes and appearances. Motivated by this, we argue that it is feasible to re-represent the object features with the common features belonging to the `sub-categories,' i.e., each category contains parts of the target objects. For example, as shown in Figure~\ref{Fig:first}, we can re-represent the dog and cat with a series of shared `sub-categories' (e.g., head, leg, body, and tail) with different weights.

\begin{figure*}[t]
\centering
    \includegraphics[width=1\linewidth]{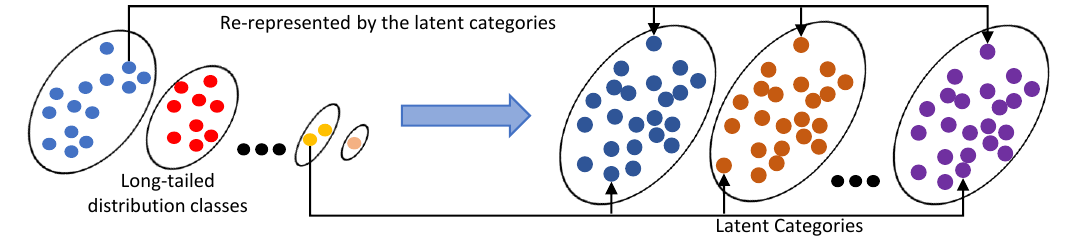}
    \caption{Our LCReg re-represent each object from the original long-tailed distribution dataset by the similarity-weighted sum of latent categories. The latent categories are shareable among the head and tailed classes and form a new balanced distributed dataset. 
    }
    \label{Fig:motivation}
\end{figure*}

To address these issues, we introduce a latent feature pool that stores common features that can be learned through backpropagation during model training. As depicted in Figure~\ref{Fig:motivation}, the latent features in the pool are class-agnostic and can be shared among all classes. To ensure that the latent features are meaningful and sufficient to represent object features, we utilize a reconstruction loss to reconstruct the original object features using the latent features, with each latent feature contributing to the reconstruction with a similarity weight. Additionally, we apply a semantic data augmentation method to the latent features in order to further increase the diversity of the training data. Our approach has several advantages due to the use of shareable latent features: 1) All object features are transferred to the shareable latent categories, making the latent features class-agnostic and no longer constrained by the imbalanced distribution. 2) Tail class objects can benefit from the diversity of head classes through the shareable latent features. 3) Tail classes can benefit from the increased diversity provided by data augmentation in the latent space, allowing for the development of latent semantic data augmentation.

In this work, we present a novel approach for long-tail recognition, referred to as Latent categories-based long-tail Recognition (LCReg). The main contribution of this method is the explicit learning of commonalities shared between head and tail classes, leading to improved feature representations. To further enhance the diversity of our training samples, we also propose the use of a semantic data augmentation method on our latent category features.

We conducted extensive experiments on a range of long-tailed recognition benchmark datasets, including CIFAR-10-LT, CIFAR-100-LT, ImageNet-LT, iNaturalist 2018, and Places-LT, to demonstrate the effectiveness of LCReg.
\section{Related Work}
\subsection{Long-Tailed Recognition.}
Most existing imbalanced classification works can be broadly classified into four categories: re-sampling, re-weighting, decoupling, and data augmentation.

\textbf{Re-sampling and Re-weighting}
Data re-sampling and loss re-weighting are common approaches for long-tailed recognition tasks. The core idea of data re-sampling is to forcibly re-balance the datasets by either under-sampling head classes \cite{DBLP:journals/corr/abs-1710-05381} or over-sampling tail classes \cite{DBLP:conf/eccv/ShenLH16,DBLP:conf/eccv/SarafianosXK18}. 

The over-sampling technique focuses on increasing the number of instances belonging to the less represented classes (also known as `tail classes') in order to address the imbalance between the more represented classes (also known as `head classes'). This has been demonstrated to be effective in previous studies such as~\cite{byrd2019effect}.
Shen\etal~\cite{shen2016relay} introduces a sampling approach called `Class-Aware Sampling (CAS)', which aims to maintain an equal probability of occurrence for each class in each batch as much as possible.
Dhruv\etal~\cite{mahajan2018exploring} developed a method for rebalancing the training data by calculating a replication factor for each image based on the distribution of labels and then repeating the images multiple times according to the replication factor. Gupta \etal~\cite{gupta2019lvis} built on this idea by proposing `Repeat Factor Sampling (RFS),' which increases the sampling frequency of images containing instances from the less represented classes (also known as `tail instances'). `Soft-balance Sampling with Hybrid Training'~\cite{peng2020large} combines traditional sampling techniques with `Class-Aware Sampling (CAS),' starting by training the detector using a conventional strategy and then introducing hyperparameters to control the amount of traditional sampling.

Unlike the over-sampling method, the under-sampling approach addresses the imbalance between the more represented classes (also known as `head classes') and the less represented classes (also known as `tail classes') by reducing the number of samples from the head classes~\cite{buda2018systematic}. 
Random under-sampling involves randomly deleting instances from the more represented classes (also known as `head classes') until they have the same number of instances as the less represented classes~\cite{buda2018systematic}. 

Over-sampling and under-sampling are popular techniques for addressing imbalanced data, but they come with some limitations. For instance, over-sampling the tail classes can lead to overfitting~\cite{chawla2002smote} and may exacerbate any errors or noise present in the tail class samples~\cite{cui2021reslt, sinha2020class}. Under-sampling, on the other hand, can result in under-learning of the head classes~\cite{sinha2020class, cui2019class,zhang2023towards} and may cause valuable data to be lost in the head classes. When dealing with extremely long-tailed data, under-sampling can often result in significant information loss due to the large difference in the amount of data between the head class and the tail class~\cite{tan2020equalization}.

In addition to re-sampling techniques, loss re-weighting approaches, such as~\cite{DBLP:journals/ida/JapkowiczS02,DBLP:conf/cvpr/TanWLLOYY20,zhao2023hierarchical,zhou2023feature}, aim to balance the loss of different classes based on the number of samples they have. However, these re-balancing methods require careful calibration of weights to avoid overfitting to the tail classes or underfitting to the head classes. Both re-sampling and re-weighting approaches can suffer from drawbacks: re-sampling often results in insufficient training of the head classes or overfitting to the tail classes, while re-weighting approaches can lead to unstable optimization during training~\cite{zhong2021improving,zhao2023weight}.
The latent feature representation has been proposed to enhance the feature representation. For instance, VQ-VAE~\cite{vq-vae}) introduced a discrete latent representation to address the problem of ``posterior collapse" and generate high-quality images.
In contrast, our proposed method, LCReg, addresses imbalanced data by transferring the unbalanced object features to shared, balanced latent categories to learn the commonalities among both the head and tail classes.

\textbf{Decoupled Training}
According to the decoupled training scheme~\cite{decouple}, training the feature extractor with the entire long-tailed dataset is beneficial, but harmful to the classifier. As a result, this two-stage approach involves first training the feature extractor and classifier on the entire long-tailed dataset, and then fine-tuning the classifier using data re-sampling to balance the weight norm of each class.
The bilateral-branch network proposes a similar decoupled training scheme~\cite{decouple} around the same time, but adds an extra classifier for fine-tuning to make the two-stage process into a single stage.
Kang \etal~\cite{kang2019decoupling} proposes a method for overcoming the problem of unbalanced data in machine learning tasks. They suggest that by separating the learning process into representation learning and classifier learning, it is possible to achieve strong long-tailed recognition. The representation learning phase can be conducted using either instance-balanced sampling or class-balanced sampling, with the results indicating that instance-balanced sampling yields the best results. 
The BAGS~\cite{li2020overcoming} also explores the idea of decoupling representation learning and classifier learning. They introduce the balanced group softmax module into the classification head of a detection framework, grouping classes according to the number of instances and executing a softmax operation group by group. This allows for the separation of classes with disparate numbers of instances, effectively balancing the classifiers in the detection framework and reducing the control of the head classes over the tail classes.
EDAL~\cite{sunevidential} proposes a novel AL framework tailored for Scene Graph Generation (SGG) task. The framework uses Evidential Deep Learning (EDL) coupled with a global relationship mining approach to estimate uncertainty and seeks diversity-based methods to alleviate context-level bias and image-level bias. 
The LPT~\cite{dong2022lpt} proposes to prompt the frozen pretrained model to adapt the long-tailed data. The prompts are divided into two groups: a shared prompt for the entire long-tailed dataset to learn general features, and group-specific prompts to gather group-specific features for samples with similar features. The two-phase training paradigm involves training the shared prompt in the first phase and optimizing group-specific prompts in the second phase with a dual sampling strategy and asymmetric Gaussian Clouded Logit loss. However, our method proposes the LCReg for long-tailed image recognition that aims to improve feature representation by learning common latent features shared by head and tail classes. 

SimCal~\cite{wang2020devil} presents a method for addressing biases in the classification head through the use of a decoupled learning scheme. The model is initially trained normally, and then a bi-level sampling scheme is used to collect class-balanced training instances through the combination of image-level and instance-level sampling. These samples are used to calibrate the classification head, improving the performance of the tail classes. To mitigate the potential negative effects of this calibration on the head classes, SimCal also introduces a Dual Head Inference architecture which selects predictions for both the tail and head classes directly from the new balanced classifier head and the original head.
In addition to the two-stage training scheme, the causal approach~\cite{tang2020long} proposes learning long-tailed datasets in an end-to-end manner by removing the negative impact of the lousy momentum effect from the causal graph.
As shown later, our proposed approach can also be used in conjunction with the decoupled training scheme.

\textbf{Data Augmentation}
Data augmentation is a common approach used to improve long-tailed recognition by creating more augmented samples to address the imbalanced distribution of data. There are several data augmentation techniques, such as generating new samples using similar samples~\cite{chawla2002smote,li2023towards,liu2020crnet,liu2022crcnet}, or other data sources~\cite{he2008adasyn,liu2021cross,liu2021fewtmm}, image flipping, scaling, rotating, and cropping. However, these techniques may not be sufficient for tail classes, which have few samples and sparse features. Mixup techniques~\cite{zhong2021improving} have been shown to help tail classes by providing enriched information from head classes. In particular, label-aware smoothing~\cite{zhong2021improving} can be used to boost classification ability during finetuning. Another approach is semantic data augmentation~\cite{isda}, which has been explored in domain adaptation~\cite{Li2021TSA} and aims to enrich the features of tail classes and create clearer decision boundaries through feature synthesis. 
The ECRT~\cite{chen2021supercharging} proposes a novel approach based on the invariance principles of causality, which allows for efficient knowledge transfer from dominant to under-represented classes using a causal data augmentation procedure.
Meta-learning has also been proposed for capturing category-wise covariance for better augmentation in long-tailed recognition~\cite{metasaug}. Our proposed method, which is built upon the work of Zhong \etal~\cite{zhong2021improving}, also uses a latent semantic augmentation loss to diversify training samples in the latent category space, providing a complementary approach to data augmentation.
\begin{figure*}[htbp]
\centering
    \includegraphics[width=1\linewidth]{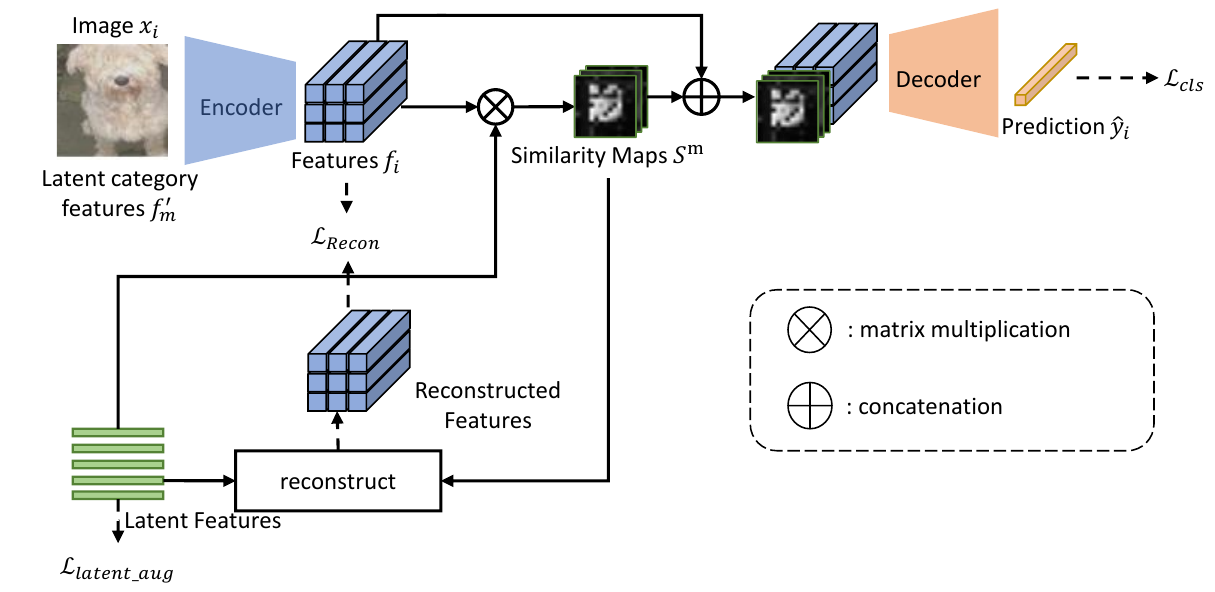}
    \vspace{-0.2cm}
    \caption{Our proposed LCReg pipeline is as follows: given an input image, we first encode its features with the Encoder. These encoded image features are then compared to the shareable latent category features, which are initially random but trainable embeddings, in order to generate similarity maps. To enhance the diversity of the latent features, we apply a latent implicit augmentation loss to the shareable latent category features. 
    Additionally, to encourage latent features that contain more object information, we reconstruct the image features using the latent features and employ a reconstruction loss. Finally, by combining the similarity maps with the original image features, we utilize a decoder to obtain the final prediction.
    }
    \label{fig:overall}
\end{figure*}
\section{Method}
In this work, we aim to optimize the performance of a classifier on a long-tail distributed dataset containing $N$ training samples with $C$ classes. Given a training sample $x_i$ with label $y_i$, our classifier uses an object feature $f_i \in \mathbb{R}^{D \times H \times W}$, generated by an encoder with parameters $\theta$, to make a prediction $\hat{y_i}$. Our goal is to minimize the distance between the prediction and the ground truth label by optimizing both the classifier and the encoder parameters $\theta$.

However, on long-tail distributed datasets, the majority of the object features $f_i$ are typically generated from head classes, leading to a bias in the classification model towards these classes and poor performance on tail classes. To address this issue, we introduce a set of class-agnostic latent features $f'$, which capture common features shared among all classes, weighted by a similarity score. In addition, we apply semantic data augmentation to these latent categories to further increase the diversity of our training samples. The overall pipeline of our proposed Latent categories-based long-tail Recognition (LCReg) method is depicted in Figure~\ref{fig:overall}.

\subsection{Latent category features}
Firstly, we introduce a set of shareable latent features $f'_{0}, f'_{1}, ...f'_{m}, ... f'_{M}$. 
Each latent feature represents a latent category that captures part of the object features and is initialized as a random learnable embedding with a dimension of $D$. These latent features can be trained through back-propagation and have a shape of $f'_{m} \in \mathbb{R}^{D \times 1}$, so all the latent feature shape is $\mathbb{R}^{D \times M}$.

We further calculate the similarity maps between latent features $f' \in \mathbb{R}^{D \times M}$ and image features $f \in \mathbb{R}^{D \times HW}$ from the image encoder, which benefits the following reconstruction process. 

\begin{equation}
    S^{m} = \sigma( \mathcal{FC}(f'_{m})^T f), 
\end{equation}
where $S^{m} \in \mathbb{R}^{1 \times H \times W}$ indicates the $m_{th}$ similarity map obtained by the $m_{th}$ encoded latent feature $\mathcal{FC}(f'_{m}) \in \mathbb{R}^{D \times 1}$ and the image feature $f$. The $\mathcal{FC}$ is a $1\times1$ convolutional layer to encode the latent features.
We normalize the map with a Sigmoid function $\sigma(\cdot)$ and then reshape the similarity map.

\subsection{Reconstruction Loss}
To encourage the latent features containing more object information, we use the latent features to reconstruct the image features $f$ by employing a reconstruction loss. Specifically, with the similarity maps ${S} \in \mathbb{R}^{M \times H \times W}$ generated by latent features, we apply a Softmax function over all the $M$ similarity maps to identify the most discriminative object parts for each latent category ${S^{m}} \in \mathbb{R}^{1 \times H \times W}$:
\begin{equation}\label{eq:get-smilarity-map}
    \hat{S}^m = \frac{\exp(S^{m})}{\sum_{k=1}^{M} \exp(S^{k})}.
\end{equation}

Then we reconstruct image features $f$ by summarizing all the latent categories with the weights from the normalized similarity maps:
\begin{equation} \label{eq:reconstruct feature}
    \hat{f} = \sum_{m=1}^{M} \mathcal{FC}(f'_{m}) \hat{S}^{m} .
\end{equation}

To compare the reconstructed features $\hat{f} \in \mathbb{R}^{D \times HW}$ and the origin features $f \in \mathbb{R}^{D \times HW}$, we calculate the correlation matrix $C_f = \hat{f}^T f$, where $C_f \in \mathbb{R}^{HW \times HW}$ and $H,W$ are the feature size. Finally, we employ a cross-entropy loss to maximize the log-likelihood of the diagonal elements of the correlation matrix $diag(C_f)$ to encourage each latent feature to learn distinct features:

\begin{equation}
    \mathcal{L}_{Recon} = -\sum_{j=1}^{HW}t_j log(\psi(diag(C_f))_j),
\end{equation}
where $j$ is the $j^{th}$ diagonal element of the correlation matrix, and $t_j \in {1,2,...,HW}$ is the pseudo ground truth of the diagonal element.
In particular, the correlation matrix $diag(C_f)$ is a matrix of size $HW * HW$. The first diagonal element of the correlation matrix is considered as the pseudo ground truth and labeled as 1, the second diagonal element is labeled as 2, and so on.
The $\psi(diag(C_f))_j$ denotes the Softmax probability for the $j^{th}$ category.

\subsection{Latent Feature Augmentation}
Data augmentation is a powerful technique that has been widely used in recognition tasks to increase training samples to reduce the over-fitting problem. 
Traditional data augmentation, such as rotation, flipping, and color-changing, are utilized to increase the training samples by changing the image itself.
In contrast to conventional data augmentation techniques, semantic data augmentation augments the semantic features by adding class-wise conditional perturbations~\cite{isda}. 
The performance of such class-conditional semantic augmentation heavily relies on the diversity of the training samples to calculate significant, meaningful co-variance matrices for perturbation sampling. However, in the long-tail recognition task, the diversity of tail classes is low due to the limited training samples. The calculated class-conditional statistics will not include sufficient meaningful semantic direction for feature augmentation, which causes negative effects on long-tailed recognition tasks. The details are shown in Section~\ref{sec. ablation-isda} and Table~\ref{Table: abalation-isda}. 

\textbf{Latent implicit semantic data augmentation.}
In contrast with ISDA~\cite{isda}, we propose to augment the latent categories to implicitly generate more training samples. To implement the semantic augmentation in the latent feature categories directly, we calculate the covariance matrices ($\boldsymbol{\Sigma}=\{\boldsymbol{\Sigma}_1, \boldsymbol{\Sigma}_2, ...,\boldsymbol{\Sigma}_M\}$) for each latent category by updating the latent features $f'_m$ at each iteration over total $M$ classes. In particular, for the $t^{th}$ training iteration, we have total $n_m^{(t)} = n_m^{(t-1)} + n^{'(t)}_m$ training samples for $m_{th}$ latent category, where the $n^{'(t)}_m$ denotes the number of training samples at the current $t^{th}$ iteration for $m_{th}$ latent category. Then we estimate the average latent feature value $\mu_m^{(t)}$ of $m_{th}$ latent category for total $t$ iteration with:

\begin{equation}
    \label{ave}
    \mu_m^{(t)} = \frac{n_m^{(t-1)}\mu_m^{(t-1)} + n^{'(t)}_m {\mu'}_m^{(t)}}
    {n_m^{(t)}},
\end{equation}
where the ${\mu'}_m^{(t)} = \frac{1}{n^{'(t)}_m} \sum_{1}^{n^{'(t)}_m}f'_m $ denotes the current average values of the latent $m_{th}$ class features at $t^{th}$ iteration. 
Then we can update the $m_{th}$ latent category covariance matrices for total $t$ training iteration with:
\begin{equation}
    \label{cv}
    \begin{split}
\Sigma_m^{(t)} 
         = \frac{n_m^{(t-1)}\Sigma_m^{(t-1)} + n^{'(t)}_m {\Sigma}_m^{'(t)}}
        {n_m^{(t)}} + \\
        \frac{n_m^{(t-1)} n^{'(t)}_m \Delta(\mu)\Delta(\mu)^T}
        {(n_m^{(t)})^2},
    \end{split}
\end{equation}
where $\Delta(\mu) = (\mu_m^{(t-1)} - {\mu'}_m^{(t)})$, and the ${\Sigma'}_m^{(t)}$ denotes the  $m_{th}$ latent category covariance matrices at current $t^{th}$ iteration.

Then, we augment the features by sampling a semantic transformation perturbation from a Gaussian distribution $\mathcal{N}(0, \lambda\boldsymbol{\Sigma}_{y'_m})$, where $\lambda$ indicates the hyperparameter of the augmentation strength and $y'_m \in {1,...,M}$ indicates the pseudo ground truth of the $M$ latent categories. In particular, we set the first latent category as the first class, the second one as the second class, and the rest in the same manner.
For each augmented latent feature $f^a_m$ we have \begin{equation}
    f^a_m \sim \mathcal{N}(f'_m, \lambda\Sigma_{y'_m}). 
\end{equation}

Furthermore, when we sample infinite times to explore all the possible meaningful perturbations in the $\mathcal{N}(0, \lambda\boldsymbol{\Sigma}_{y'_m})$, there is an upper bound of the cross-entropy loss~\cite{isda} on all the augmented features over $N$ training samples:
\begin{align}
    \label{Eq: loss-embedding}
    \mathcal{L}_{latent\_aug} &= \sum_{i=1}^{N} L_{ \infty}(f(\boldsymbol{x_i};\theta),y'_m; \boldsymbol{\Sigma}) \\ \notag
& = \frac{1}{N} \sum_{i=1}^{N} log(\sum_{j=1}^{M} e^{ z_j
})
\end{align}
\begin{equation}
\begin{split}
z_j = (\boldmath{w}^{T}_{j} - \boldmath{w}^{T}_{y'_{m}})f^a_m + (b_{j} - b_{y'_{m}}) + \\ \frac{\lambda}{2}(\boldmath{w}^{T}_{j} - \boldmath{w}^{T}_{y'_{m}})\Sigma_{y'_m}(\boldmath{w}_{j} - \boldmath{w}_{y'_{m}}),
\end{split}
\end{equation}
where $\theta$ indicates the encoder parameters for the latent category features. 
$\boldmath{w}$ and $\boldmath{b}$ are the weight and biases corresponding to the a $1\times1$ convolution layer $\mathcal{FC}$ motioned above. Following ISDA~\cite{isda}, we let $\lambda  = (t/T)\!\times\!\lambda_0$ to reduce the augmentation impact in the beginning of the training stage, where $T$ indicates the total iteration. 

With the augmented latent category features, we are able to increase the diversity of training samples by reconstructing the augmented latent features back to the image features $f$ with the reconstruction loss $\mathcal{L}_{Recon}$.

\subsection{Training Process}
We adopt decoupled training for the long-tailed task as in \cite{zhong2021improving}. Specifically, in the first stage of the training process, our training objective includes the reconstruction loss $\mathcal{L}_{Recon}$ which is applied on the latent category features, a latent augmentation loss $\mathcal{L}_{latent\_aug}$ that augments the latent features, and a cross-entropy classification loss which is applied on final prediction $\hat{y_i}$ generated with the decoder.
We optimize the network parameter by combining all the losses:
\begin{equation}
\label{eqn:loss}
\mathcal{L} =\alpha \mathcal{L}_{latent\_aug}  + \beta\mathcal{L}_{Recon} + \gamma\mathcal{L}_{cls}, 
\end{equation}
where ${L}_{cls}$ indicates the final classification loss (CE loss) between the ground truth $y$ and the prediction $\hat{y_i}$.  $\alpha$, $\beta$, and $\gamma$ are the trade-off parameters, which have been set to 0.1, 0.1, and 1, respectively. 
In the second stage of training, following~\cite{zhong2021improving}, we finetune the network. 
\section{Experiments}
\subsection{Implementation Details}
We follow the training pipeline described in previous works~\cite{zhong2021improving,bbn} to conduct experiments on five datasets: CIFAR-10-LT, CIFAR-100-LT, ImageNet-LT, iNaturalist 2018, and Places-LT. We use the SGD optimizer and apply data augmentation techniques such as random scaling, cropping, and flipping during training. Unless otherwise stated, we use a batch size of 128 for all experiments.

\subsection{Dataset}
\noindent\textbf{CIFAR-10-LT and CIFAR-100-LT.}
We conduct experiments on the long-tailed versions of the CIFAR datasets, as described in~\cite{cao2019learning}. The CIFAR-10 and CIFAR-100 datasets consist of 50,000 and 10,000 training and validation images, respectively, across 10 and 100 categories. To create a long-tailed dataset, we discard some of the training samples and rearrange the remaining ones to create an imbalance factor (IF) $ = N_{max}/N_{min}$, where $N_{max}$ and $N_{min}$ are the numbers of training samples for the largest and smallest classes, respectively. Following previous works~\cite{cao2019learning,zhong2021improving,liu2023harmonizing}, we conduct experiments on the CIFAR-LT datasets with IF values of 10, 50, and 100.

\noindent\textbf{ImageNet-LT.} Liu~\etal~\cite{liu2019large} propose the ImageNet-LT dataset, which contains 115,846 training images and 50,000 validation images, including 1000 categories, with the imbalance factor(IF) of 1280/5. This dataset is a subset of ImageNet~\cite{imagenet}. They follow the Pareto distribution with power value = 6 to sample the images and rearrange to a new unbalanced dataset. 

\noindent\textbf{iNaturalist 2018.} iNaturalist 2018~\cite{van2018inaturalist} is a large-scale dataset collected from the real world, whose distribution is extremely unbalanced. It contains 435,713 images for 8142 categories with an imbalanced factor(IF) of 1000/2. 

\noindent\textbf{Places-LT.} Places-LT is a long-tailed distribution dataset generated from the large-scale scene classification dataset Places~\cite{places}. It consists of 184.5K images for 365 categories with an imbalanced factor(IF) of 4980/5. 

\subsection{Comparisons with State-of-the-art methods}
\noindent\textbf{Experiments on CIFAR-LT.} Following previous works~\cite{zhong2021improving,tang2020long,ldam,bbn,liu2020guided}, we conduct experiments on the CIFAR-10-LT and CIFAR-100-LT datasets with imbalance factors of 10, 50, and 100. The latent categories are set to 40 and 50 for CIFAR-10-LT and CIFAR-100-LT, respectively. As shown in Table~\ref{tab:cifar}, our proposed method outperforms all previous methods.

\begin{table}[t]
\centering
		\centering
		\resizebox{0.6\linewidth}{!}{
		\begin{tabular}{l|ccc|ccc}
			\toprule[1.5pt]
			{\multirow{2.6}*{\textbf{Method}}}	 &\multicolumn{3}{c|}{CIFAR-10-LT} & \multicolumn{3}{c}{CIFAR-100-LT} \\ 
			\cmidrule(lr){2-4}\cmidrule(l){5-7}
			& 100 & 50 &  10 & 100 & 50 & 10 \\ 
			\hline
			
			CE (Cross Entropy) & 70.4  & 74.8  & 86.4  & 38.4  & 43.9  &  55.8   \\
			mixup~\cite{mixup} & 73.1  & 77.8 &  87.1  & 39.6  & 45.0  &  58.2   \\
			LDAM+DRW~\cite{ldam} & 77.1  & 81.1  &  88.4  & 42.1 & 46.7  &  58.8    \\

			BBN{\scriptsize{(include mixup)}}~\cite{bbn} & 79.9  & 82.2  &  88.4  & 42.6  & 47.1  &  59.2  \\
			Remix+DRW~\cite{remix} & 79.8  & -  &  89.1  & 46.8  & -  &  61.3  \\
			
			{MiSLAS~\cite{zhong2021improving}}  & {82.1}   & {85.7} &  {90.0}  & {47.0}  & {52.3}   &  {63.2}  \\
			MetaSAug CE\cite{metasaug} & 80.5 & 84.0 & 89.4 & 46.9 & 51.9 & 61.7 \\

MetaSAug LDAM~\cite{metasaug} & 80.7 & 84.4 & 89.7 & {48.0} & 52.2 & 61.2 \\

PaCo~\cite{PaCo} & - & - & - & \textbf{52.0} & \textbf{56.0} &64.2   \\

			\hline   

			{Ours}  & \textbf{83.1}   & \textbf{86.5} &  \textbf{91.2}  & 47.6  & {53.1}   &  \textbf{64.2} \\ 
			\bottomrule[1.5pt]     
		\end{tabular}
		}
		\caption{The top-1 accuracy (in \%) for ResNet-32 based models trained on the CIFAR-10-LT and CIFAR-100-LT datasets.}
		\label{tab:cifar}
	\end{table}

\noindent\textbf{Experiments on large-scale datasets.} 
We further evaluate the effectiveness of our method on the large-scale, imbalanced datasets ImageNet-LT, iNaturalist 2018, and Places-LT. The latent category numbers are set to 100 for ImageNet-LT and 200 for iNaturalist 2018, and 100 for the Places-LT dataset. As shown in Tables~\ref{Table: soa_imagenet}, \ref{Table: soa_inat}, and \ref{Table: soa_places}, our proposed method improves the baseline methods by leveraging the shared commonalities between head and tail classes and employing semantic data augmentation on latent category features, achieving comparable performance to previous state-of-the-art methods on all large-scale datasets.
\begin{table}[t]
\centering
\small
\resizebox{0.4\linewidth}{!}{
\begin{tabular}{p{4.3cm}|p{1.6cm}<{\centering}}
					\toprule[1.5pt]
					\textbf{Method}              & ResNet-50     \\ \hline
					CE & 44.6         \\
					CE+DRW~\cite{ldam} & 48.5         \\
					Focal+DRW~\cite{lin2017focal}  & 47.9 \\
					LDAM+DRW~\cite{ldam}    & 48.8 \\ 
					NCM \cite{decouple}  & 44.3  \\
$\tau$-norm \cite{decouple} & 46.7   \\
cRT \cite{decouple} & 47.3     \\
LWS \cite{decouple} & 47.7     \\
{MiSLAS~\cite{zhong2021improving}}      & {52.7} \\
MetaSAug CE \cite{metasaug} & 47.4\\

PaCo*~\cite{PaCo} & 51.0   \\

PaCo~\cite{PaCo} & \textbf{57.0}   \\

RIDE(2 experts)~\cite{ride}  &54.4 \\
						\hline
							Ours  &  {55.3}      \\
					\bottomrule[1.5pt]
					
\end{tabular}
}
\caption{The top-1 accuracy (in \%) for the ResNet-50 based models trained on ImageNet-LT. * denotes without RandAugment method. 
}
\label{Table: soa_imagenet}
\end{table}
\begin{table}[t]
\centering
\small
\resizebox{0.4\linewidth}{!}{
\begin{tabular}{p{4.3cm}|p{1.6cm}<{\centering}}
					\toprule[1.5pt]
					\textbf{Method}              & ResNet-50 \\ \hline
					CB-Focal~\cite{effnum}  & 61.1\\
					LDAM+DRW~\cite{ldam}  &68.0 \\ 
					OLTR~\cite{liu2019large}   & 63.9 \\
					cRT~\cite{decouple}  & 65.2 \\
					${\tau}$-norm~\cite{decouple}  & 65.6 \\
					LWS~\cite{decouple} & 65.9 \\
					BBN{\scriptsize(include mixup)}~\cite{bbn} & 69.6 \\
					Remix+DRW~\cite{remix} & 70.5  \\ 
					{MiSLAS~\cite{zhong2021improving}}  & {71.6} \\ 
					MetaSAug CE \cite{metasaug} & 68.8\\

PaCo~\cite{PaCo} & \textbf{73.2}   \\

RIDE(2 experts)~\cite{ride}  &71.4 \\

					\hline
						Ours  & {72.6}      \\ 
					\bottomrule[1.5pt]
\end{tabular}
}
\caption{The top-1 accuracy (in \%) for the ResNet-50 based models trained on iNaturalist 2018.
}
\label{Table: soa_inat}
\end{table}
\begin{table}[t]
\centering
\small
\resizebox{0.35\linewidth}{!}{
\begin{tabular}{l|c}
\toprule[1.5pt]
					\textbf{Method}              & ResNet-152     \\ \hline  
					Range Loss~\cite{zhang2017range}      & 35.1 \\
					FSLwF~\cite{gidaris2018dynamic} & 34.9 \\ 
					OLTR~\cite{liu2019large} & 35.9 \\
					OLTR+LFME~\cite{xiang2020learning} & 36.2 \\
PaCo~\cite{PaCo} & \textbf{41.2}  \\

						\hline
					Ours  &    {40.2}         \\ 
					\bottomrule[1.5pt]
\end{tabular}
}
\caption{The top-1 accuracy (in \%) for the ResNet-152 based models trained on Places-LT.
}
\label{Table: soa_places}
\end{table}

\subsection{Ablation Studies}
\noindent\textbf{Number of the latent categories.}
We conduct experiments to analyze the impact of the number of latent categories on performance for different datasets. As shown in Table~\ref{Table: abalation-emb}, we experiment with both small and large-scale datasets to explore the effectiveness of the number of latent categories.
For larger datasets, which have more training samples and classes, we suggest using more latent categories to better represent the original image features and achieve better performance. However, simply increasing the number of latent categories does not always lead to improved performance. For example, 40 categories yield the best performance on the CIFAR-10-LT dataset, while further increasing the number of categories leads to a rapid decline in performance.
We speculate that having too many latent categories may result in the object features being split too finely, resulting in a loss of meaningful parts.
Specifically, datasets similar in size and categories to CIFAR-10-LT or CIFAR-100-LT tend to exhibit better performance when the latent categories are set to approximately 20-60. Similarly, datasets comparable to ImageNet-LT demonstrate enhanced performance when the latent categories are set to around 100-300. For datasets similar to iNaturalist in terms of size and categories, setting the latent categories to a value larger than 200 yields improved performance.

\begin{table*}[t]
\centering
\resizebox{0.9\linewidth}{!}{
\begin{tabular}{l|c|c|c|c|c|l|c|c|c|c}
\toprule[1.5pt]
Dataset  & \multicolumn{5}{c}{Number of latent class} &  \multicolumn{1}{|l}{Dataset}         & \multicolumn{4}{|c}{Number of latent class} \\
\hline
         & 20         & 30        & 40        & 50        & 60        &                 & 20            & 60           & 100          & 200          \\  \hline
CIFAR-10-LT  & 81.9       & 82.4      & 83.1      & 82.5      & 79.6      & ImageNet-LT     & 54.5          & 55.0         & 55.3         & 55.2            \\
CIFAR-100-LT & 47.1       & 47.2      & 47.4      & 47.6      & 46.1      & iNaturalist 2018 & -             & 71.6            & 71.6         & 72.6        \\
\bottomrule[1.5pt]
\end{tabular}
}
\caption{The results of ablation studies on the effectiveness of the number of latent categories for long-tailed image recognition tasks. We conduct experiments on both small datasets (CIFAR-10-LT and CIFAR-100-LT with imbalance factor (IF) 100) and large datasets (ImageNet-LT and iNaturalist 2018).
The results show that as the size of the dataset increases (with more training samples and classes), a larger number of latent categories is generally required to achieve better performance. However, it is also noted that continuously increasing the number of latent categories beyond a certain point may not necessarily lead to further improvement and may even result in a decrease in performance.
}
\label{Table: abalation-emb}
\end{table*}

\noindent\textbf{Performance on different splits of classes.}
We also report the classification accuracy for classes with a large number of images (more than 100 per class), a medium number of images (20 to 100 per class), and a small number of images (less than 20 per class). In particular, we set the number of latent categories to 40 for CIFAR-10-LT, 50 for CIFAR-100-LT, 100 for ImageNet-LT, and 200 for iNaturalist 2018.
As shown in Table~\ref{many-medium-few}, our method consistently outperforms all other methods by a large margin for all classes on all datasets.
\begin{table*}[]
\centering
\resizebox{0.63\linewidth}{!}{
\begin{tabular}{l|l|c|c|c} 
\toprule[1.5pt]
Dataset                           & Methods      & Many   & Medium & Few    \\ \hline

\multirow{2}{*}{CIFAR10-LT IF 100}       & Ours$^*$     & {90.9} & {80.8} & {73.7} \\
                                  & \cellcolor[HTML]{9B9B9B}Ours         & \cellcolor[HTML]{9B9B9B}\textbf{92.6} & \cellcolor[HTML]{9B9B9B}\textbf{81.5} & \cellcolor[HTML]{9B9B9B}\textbf{75.4} \\   \hline

\multirow{6}{*}{CIFAR100-LT IF 100}      & OLTR~\cite{liu2019large}         & 61.8   & 41.4   & 17.6   \\ 
                                  & LDAM + DRW~\cite{cao2019learning}   & 61.5   & 41.7   & 20.2   \\
                                  & ${\tau}$-norm~\cite{decouple}       & 65.7   & 43.6   & 17.3   \\
                                  & cRT~\cite{decouple}          & 64.0   & 44.8   & 18.1   \\
                                  & Ours$^*$     & 63.1       & {48.6}       &  {25.0}      \\
                                  & \cellcolor[HTML]{9B9B9B}Ours         & \cellcolor[HTML]{9B9B9B}\textbf{64.2} & \cellcolor[HTML]{9B9B9B}\textbf{49.2}  & \cellcolor[HTML]{9B9B9B}\textbf{25.4}  \\  \hline
\multirow{6}{*}{ImageNet-LT}      & cRT~\cite{decouple}          & 62.5   & 47.4   & 29.5   \\
                                  & LWS~\cite{decouple}          & 61.8   & 48.6   & 33.5   \\
                                  & Ours$^*$     & 61.7   & 51.2   & 35.6   \\
                                  & \cellcolor[HTML]{9B9B9B}Ours         & \cellcolor[HTML]{9B9B9B}\textbf{66.1}   & \cellcolor[HTML]{9B9B9B}\textbf{52.8}   & \cellcolor[HTML]{9B9B9B}\textbf{36.2}   \\  \hline
\multirow{5}{*}{iNaturalist 2018} & cRT~\cite{decouple}          & 73.2   & 68.8   & 66.1   \\
                                  & ${\tau}$-norm~\cite{decouple} & 71.1   & 68.9   & 69.3   \\
                                  & LWS~\cite{decouple}          & 71.0   & 69.8   & 68.8   \\
                                  & Ours$^*$     & 73.2   & 72.4   & 70.4   \\
                                  & \cellcolor[HTML]{9B9B9B}Ours         & \cellcolor[HTML]{9B9B9B}\textbf{73.8} & \cellcolor[HTML]{9B9B9B}\textbf{73.4} & \cellcolor[HTML]{9B9B9B}\textbf{71.5} \\  \hline
                                  
\end{tabular}
}
\caption{
We evaluate the accuracy of our proposed methods on three different splits of classes: Many, Medium, and Few. To validate the effectiveness of our approach, we conduct experiments on a variety of datasets, including small-scale datasets such as CIFAR10-LT and CIFAR100-LT with IF 100, as well as large-scale datasets like ImageNet-LT and iNaturalist 2018.
For comparison, we also report the results of our baseline approach, indicated as``Ours$^*$," which does not incorporate latent category features or the reconstruction loss $\mathcal{L}{Recon}$ and latent augmentation loss $\mathcal{L}{latent_aug}$. This allows us to assess the impact of these additional components on the overall performance of our method.
}
\label{many-medium-few}
\end{table*}

\noindent\textbf{Effect of each component.}
We investigate the contribution of each component of our proposed method - the latent categories, the latent augmentation loss, and the latent reconstruction loss - by conducting ablation experiments on both small and large-scale datasets. Specifically, we choose an imbalance factor (IF) of 100 and set the number of latent categories to 40 for CIFAR-10-LT and 50 for CIFAR-100-LT.
For the experiments on the large challenge datasets (iNaturalist 2018), we set the number of latent categories to 100 but used a smaller training batch size of 16 due to resource constraints. As shown in Table~\ref{Table: abalation-losses}, adding our proposed latent categories alone significantly improves the performance of the baseline method on all datasets. The performance is further improved by applying the latent augmentation loss and latent reconstruction loss.

\begin{table}[t]
\centering
\small
\resizebox{1\linewidth}{!}{
\begin{tabular}{c|c|c|c|c|c|c|c|c|c}
\toprule[1.5pt]
\multicolumn{3}{c}{Components}              & \multicolumn{3}{|c}{CIFAR-10-LT}  & \multicolumn{3}{|c}{CIFAR-100-LT} & \multicolumn{1}{|c}{iNaturalist 2018}                \\ \hline
latent category & latent aug & latent recon &100 &50 &10 & 100 &50 &10 & - \\ \hline
                &            &              &   82.1   &85.7 &90.0    &    47.0   &52.3 &63.2             &68.9            \\ 
\checkmark      &            &              &   82.2   &85.8 &90.7    &    47.2   &52.6 &63.9             &69.4            \\ 
\checkmark      & \checkmark &              &   82.5   &86.0 &91.0   &     47.4   &53.0 &64.1            &69.8            \\ 
\checkmark      &            & \checkmark   &   83.0   &86.2 &91.1    &    47.3   &52.5 & 64.0            &70.0            \\ 
\checkmark      &\checkmark  & \checkmark   &   \textbf{83.1}   &\textbf{86.5} &\textbf{91.2}    &    \textbf{47.6}   &\textbf{53.1} & \textbf{64.2}            &\textbf{70.5}              \\
\bottomrule[1.5pt]
\end{tabular} 
}
\caption{The results of ablation studies on the effectiveness of each component of our proposed method for long-tailed image recognition tasks. We conduct experiments on both small datasets (CIFAR-10-LT and CIFAR-100-LT with imbalance factor (IF) 100, 50, and 10) and a large dataset (iNaturalist 2018).
The results show that each of our proposed components (utilizing latent categories, latent augmentation loss, and latent reconstruction loss) individually improves the performance of the baseline (without any of the proposed components) on all datasets. This demonstrates the effectiveness of each component in improving the performance of long-tailed image recognition tasks.
}
\label{Table: abalation-losses}
\end{table}

\begin{figure*}[t]
\centering
    \includegraphics[width=1\linewidth]{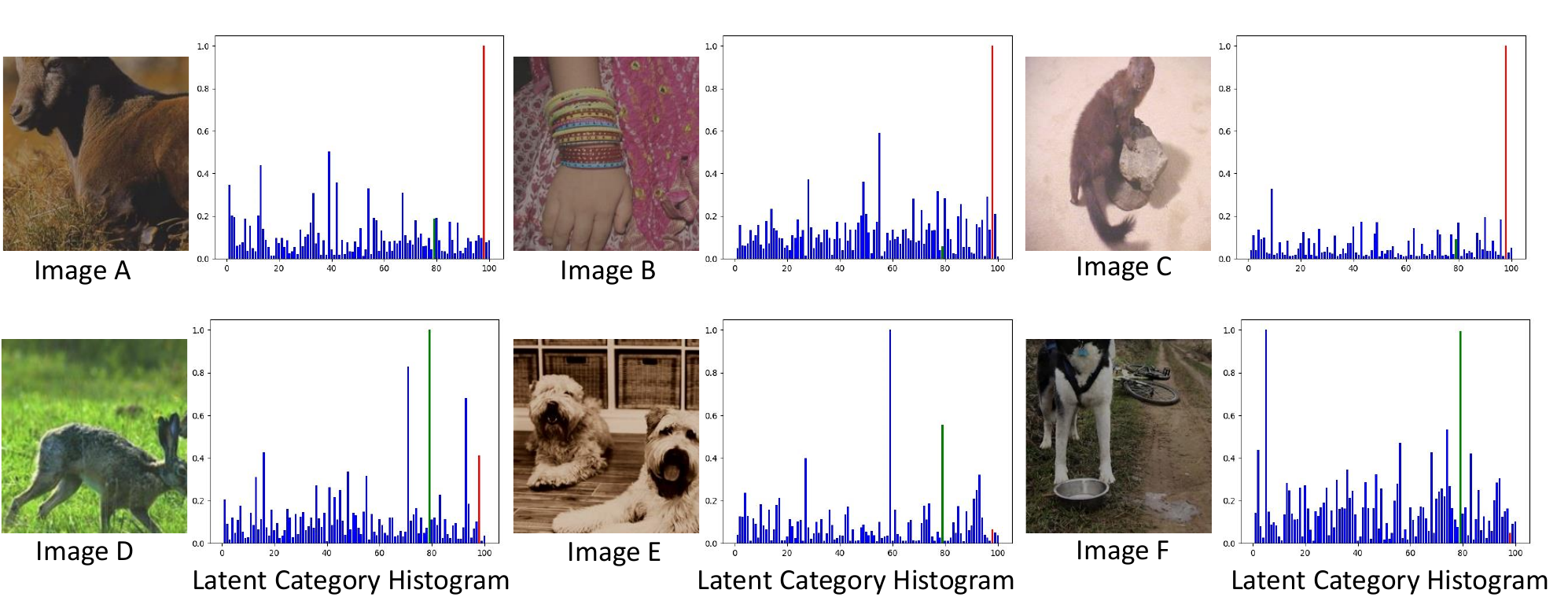}
    \caption{The weight histogram of latent categories contributing to the reconstruction of image features for a sample of images from the ImageNet-LT dataset. As depicted in the figure, the $79^{th}$ latent category (green) is highlighted by the hare' (Image D), dogs' (Image E and F), which may be because of containing similar shapes of limbs. Additionally, the cow' (Image A), human arm' (Image B), and `fisher' (Image C) share some commonalities captured by the $98^{th}$ latent category (red). It is important to note that our proposed method aims to learn commonalities between images belonging to latent classes, which are not necessarily denoted by appearances from a human perspective. A common characteristic can be any characteristic of an object, such as color, structure, or shape.}
    \label{Fig:visualization}
\end{figure*}

\noindent\textbf{Visualization of the latent categories.}
As shown in Figure~\ref{Fig:visualization}, we visualize the latent category histogram for the ImageNet-LT dataset with 100 latent categories. We reconstruct the image features using the latent categories, with each latent category contributing a normalized similarity weight generated by equation~\ref{eq:get-smilarity-map}. As shown in the figure, the 79th latent category (green) is highlighted for the hare' and dogs' (Images E and F), maybe due to their similar limb patterns. Additionally, the cow', human arm', and `fisher' also share some commonalities captured by the 98th latent category (red). 

\begin{figure*}[t]
\centering
    \includegraphics[width=1\linewidth]{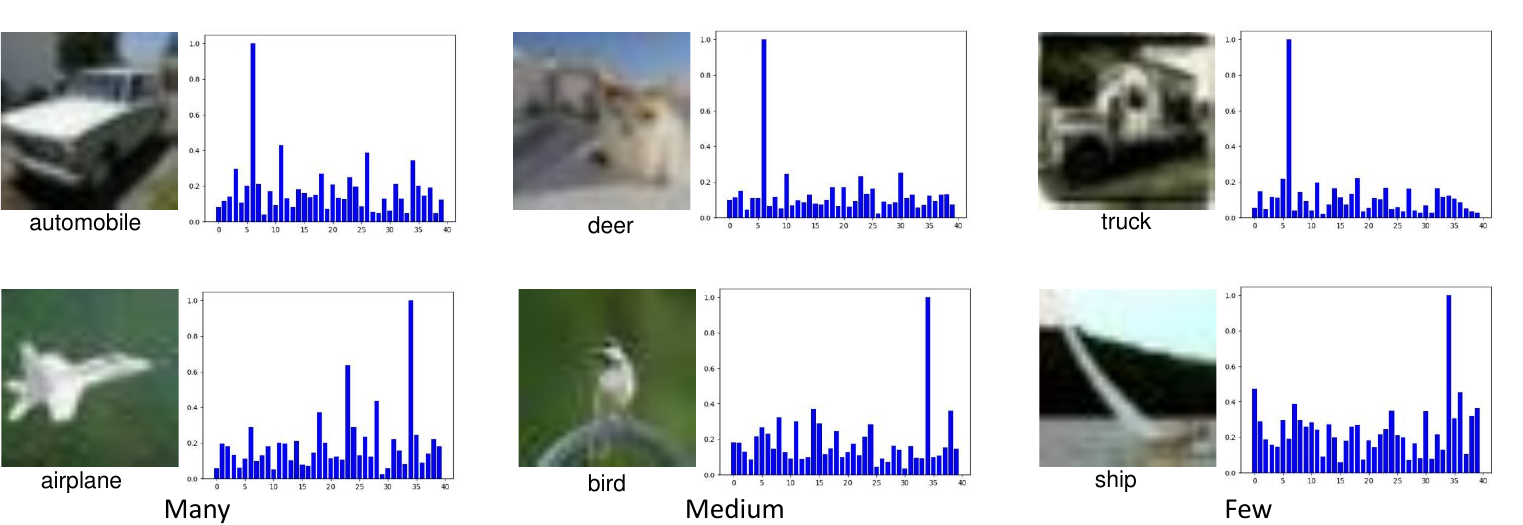}
    \caption{The weight histogram of latent categories contributing to the reconstruction of image features for a sample of images from the CIFAR-10 dataset. The figure illustrates that objects from many classes, medium classes, and few classes still share some commonalities. These results suggest that our approach is able to capture shared features across different classes.}
    \label{Fig:visualization_cifar}
\end{figure*}

As shown in Figure~\ref{Fig:visualization_cifar}, we have included the examples from the CIFAR-10-LT dataset. The figure illustrates that objects from many classes, medium classes, and few classes share some commonalities.
To further analyse the histogram of objects from different classes, we have calculated their KL divergence loss in Table~\ref{tab:similarity}. As the table demonstrates, the KL divergence loss between objects of different classes can still be small, such as the objects of the automobile class (from many classes) to the ship class (from few classes). This validates our assumption of shareable commonalities.
 \begin{table}[h]
\centering
\begin{tabular}{c|c|c|c|c|c|c}
\toprule[1.5pt]
& automobile & deer & truck & airplane & bird & ship \\ \hline
automobile & 0.00 & 4.39 & 4.90 & 1.58 & 2.04 & 0.52 \\
deer & -0.30 & 0.00 & 1.85 & 0.44 & 0.76 & -0.87 \\
truck & -0.99 & 0.35 & 0.00 & -0.08 & -0.03 & -0.80 \\ 
airplane & 2.89 & 6.54 & 7.76 & 0.00 & 1.71 & 0.07 \\
bird & 2.63 & 5.94 & 6.60 & 1.14 & 0.00 & -0.66 \\
ship & 6.37 & 9.95 & 11.92 & 5.01 & 4.71 & 0.00 \\ \bottomrule[1.5pt]
\end{tabular}
\caption{The KL divergence loss.}
\label{tab:similarity}
\end{table}

\noindent\textbf{The effectiveness of hyper-parameter.}
As shown in Table~\ref{Table: abalation-parameter}, we investigate the impact of the hyperparameters $\alpha$, $\beta$, and $\gamma$ in Equation~\ref{eqn:loss} on the performance of our method. The results in the table demonstrate that our method is relatively insensitive to these hyperparameters. This suggests that our method is robust and can achieve stable performance across a wide range of hyperparameter values.
\begin{table}[t]
\centering
\resizebox{0.5\linewidth}{!}{
\begin{tabular}{c|c|c|c|c|c|c|c|c}
\toprule[1.5pt]
  & & & \multicolumn{3}{c}{CIFAR-10-LT}      & \multicolumn{3}{|c}{CIFAR-100-LT} \\
\hline
       $\alpha$  & $\beta$ &$\gamma$  & 100         & 50        & 10                      & 100         & 50        & 10                  \\  \hline
1& 1 & 1& {82.0}   & {85.6}    &{91.0}   & {47.0}  & {52.8}   &  {63.8}        \\
1& 0.1 & 1 & {82.2}   & {85.8}    &{89.8}   & {47.3}  & {52.6}   &  {63.7}        \\      
1& 1 & 0.1    & {82.4}   & {85.8}    &\textbf{91.1}   & {47.1}  & {52.7}   &  {63.9}        \\

0.1& 0.1 & 1    & \textbf{82.5}   & \textbf{86.0}    &{91.0}   & \textbf{47.4}  & \textbf{53.0}   &  \textbf{64.1}        \\
\bottomrule[1.5pt]
\end{tabular} 
}
\caption{The results of ablation studies on the parameter selection for our proposed method on long-tailed image recognition tasks. We report the top-1 accuracy (\%) for different values of the parameters $\alpha, \beta$ and $\gamma$ on various datasets.
The results show that our method is relatively insensitive to the choice of these parameters, achieving good performance across a wide range of parameter values. This demonstrates the robustness and flexibility of our proposed method in handling long-tailed image recognition tasks.
}
\label{Table: abalation-parameter}
\end{table}

\noindent\textbf{Latent augmentation vs. ISDA} \label{sec. ablation-isda}
As shown in Table~\ref{Table: abalation-isda}, we directly apply the ISDA method~\cite{isda} to the original class features and conduct experiments on CIFAR-10-LT and CIFAR-100-LT with different imbalance impacts.
Our proposed latent augmentation method, which augments the features within the latent categories, significantly improves the performance on all long-tail recognition datasets, compared to directly using the unbalanced features. 
This demonstrates the effectiveness of our latent augmentation method in improving the performance of long-tailed image recognition tasks.
Specifically, we set the number of latent categories to 40 for CIFAR-10-LT and 50 for CIFAR-100-LT.
\begin{table}[t]
\centering
\resizebox{0.5\linewidth}{!}{
\begin{tabular}{l|c|c|c|c|c|c}
\toprule[1.5pt]
  & \multicolumn{3}{c}{CIFAR-10-LT}      & \multicolumn{3}{|c}{CIFAR-100-LT} \\
\hline
      Methods   & 100         & 50        & 10                      & 100         & 50        & 10                  \\  \hline
Baseline & {82.1}   & {85.7} &  {90.0}                  & {47.0}  & {52.3}   &  {63.2}        \\
+ ISDA & {79.8}   & {82.7} &  {87.8}                    & {43.5}  & {47.8}   &  {57.7}         \\
+ $L_{Aug}$    & \textbf{82.5}   & \textbf{86.0}    &\textbf{91.0}   & \textbf{47.4}  & \textbf{53.0}   &  \textbf{64.1}        \\
\bottomrule[1.5pt]
\end{tabular} 
}
\caption{Ablation studies comparing the normal feature augmentation with ISDA to the latent feature augmentation. The top-1 accuracy (\%) is reported for various datasets. The results show that using the latent augmentation method on the latent category features, denoted as $L_{Aug}$, significantly improves the performance compared to using normal feature augmentation with ISDA on the original class features. This suggests that augmenting the latent category features, which capture commonalities among different classes, is more effective than augmenting the original class features directly.
}
\label{Table: abalation-isda}
\end{table}
\section{Conclusion}
In this work, we have proposed a novel approach for addressing the challenges of long-tailed image recognition, called latent category-based long-tail recognition (LCReg). This method aims to increase the diversity of training samples for long-tailed recognition tasks by mining and augmenting common features among head and tail classes. To achieve this, we introduce latent category features, which are class-agnostic and can be shared among all classes. These features are learned through backpropagation during model training and are used to reconstruct the original object features using a reconstruction loss. In addition, we apply a semantic data augmentation method to these latent features to further increase their diversity. Our approach has several advantages over traditional methods, including the ability to handle small-data learning problems and the use of class-agnostic features that can be shared among all classes. Our experimental results on several long-tailed recognition benchmarks demonstrate the effectiveness of our method.

\section*{Acknowledgements} This research is supported by the National Research Foundation, Singapore under its AI Singapore Programme (AISG Award No: AISG-RP-2018-003), the Ministry of Education, Singapore, under its Academic Research Fund Tier 1 (RG95/20). This research is partly supported by the Agency for Science, Technology and Research (A*STAR) under its AME Programmatic Funds (Grant No. A20H6b0151).

\bibliographystyle{elsarticle-num}
\bibliography{egbib}
\end{document}